\documentclass[letterpaper]{article} % DO NOT CHANGE THIS
\usepackage{aaai24}  % DO NOT CHANGE THIS
\usepackage{times}  % DO NOT CHANGE THIS
\usepackage{helvet}  % DO NOT CHANGE THIS
\usepackage{courier}  % DO NOT CHANGE THIS
\usepackage[hyphens]{url}  % DO NOT CHANGE THIS
\usepackage{graphicx} % DO NOT CHANGE THIS
\usepackage{natbib}  % DO NOT CHANGE THIS AND DO NOT ADD ANY OPTIONS TO IT
\usepackage{caption} % DO NOT CHANGE THIS AND DO NOT ADD ANY OPTIONS TO IT
\usepackage{algorithm}
\usepackage{algorithmic}

\usepackage{newfloat}
\usepackage{listings}
\DeclareCaptionStyle{ruled}{labelfont=normalfont,labelsep=colon,strut=off} % DO NOT CHANGE THIS
\floatstyle{ruled}
\newfloat{listing}{tb}{lst}{}
\floatname{listing}{Listing}
\usepackage{nth}
\usepackage{booktabs}
\usepackage{multirow}
\usepackage{amssymb}
\usepackage{array}
\usepackage{makecell}
\usepackage{colortbl}
\usepackage[table, dvipsnames]{xcolor}
\usepackage{amsmath}
\usepackage{dsfont}

\title{EDA: Evolving and Distinct Anchors for Multimodal Motion Prediction}
\author{
    Longzhong Lin\textsuperscript{\rm 1,2},
    Xuewu Lin\textsuperscript{\rm 2},
    Tianwei Lin\textsuperscript{\rm 2},
    Lichao Huang\textsuperscript{\rm 2},
    Rong Xiong\textsuperscript{\rm 1},
    Yue Wang\textsuperscript{\rm 1}\thanks{Corresponding author.}
}
\affiliations{
    \textsuperscript{\rm 1}Zhejiang University,
    \textsuperscript{\rm 2}Horizon Robotics
    linlongzhong2000@zju.edu.cn, \{xuewu.lin, tianwei.lin, lichao.huang\}@horizon.cc, \{rxiong, ywang24\}@zju.edu.cn
}

\usepackage{bibentry}
\begin{document}

\maketitle

\begin{abstract}
Motion prediction is a crucial task in autonomous driving, and one of its major challenges lands in the multimodality of future behaviors.
Many successful works have utilized mixture models which require identification of positive mixture components, and correspondingly fall into two main lines: prediction-based and anchor-based matching.
The prediction clustering phenomenon in prediction-based matching makes it difficult to pick representative trajectories for downstream tasks, while the anchor-based matching suffers from a limited regression capability.
In this paper, we introduce a novel paradigm, named Evolving and Distinct Anchors (EDA), to define the positive and negative components for multimodal motion prediction based on mixture models.
We enable anchors to evolve and redistribute themselves under specific scenes for an enlarged regression capacity.
Furthermore, we select distinct anchors before matching them with the ground truth, which results in impressive scoring performance.
Our approach enhances all metrics compared to the baseline MTR, particularly with a notable relative reduction of 13.5\% in Miss Rate, resulting in state-of-the-art performance on the Waymo Open Motion Dataset.
Code is available at https://github.com/Longzhong-Lin/EDA.
\end{abstract}

\section{Introduction}

\begin{figure}[tb]
\centering
\includegraphics[width=0.99\columnwidth]{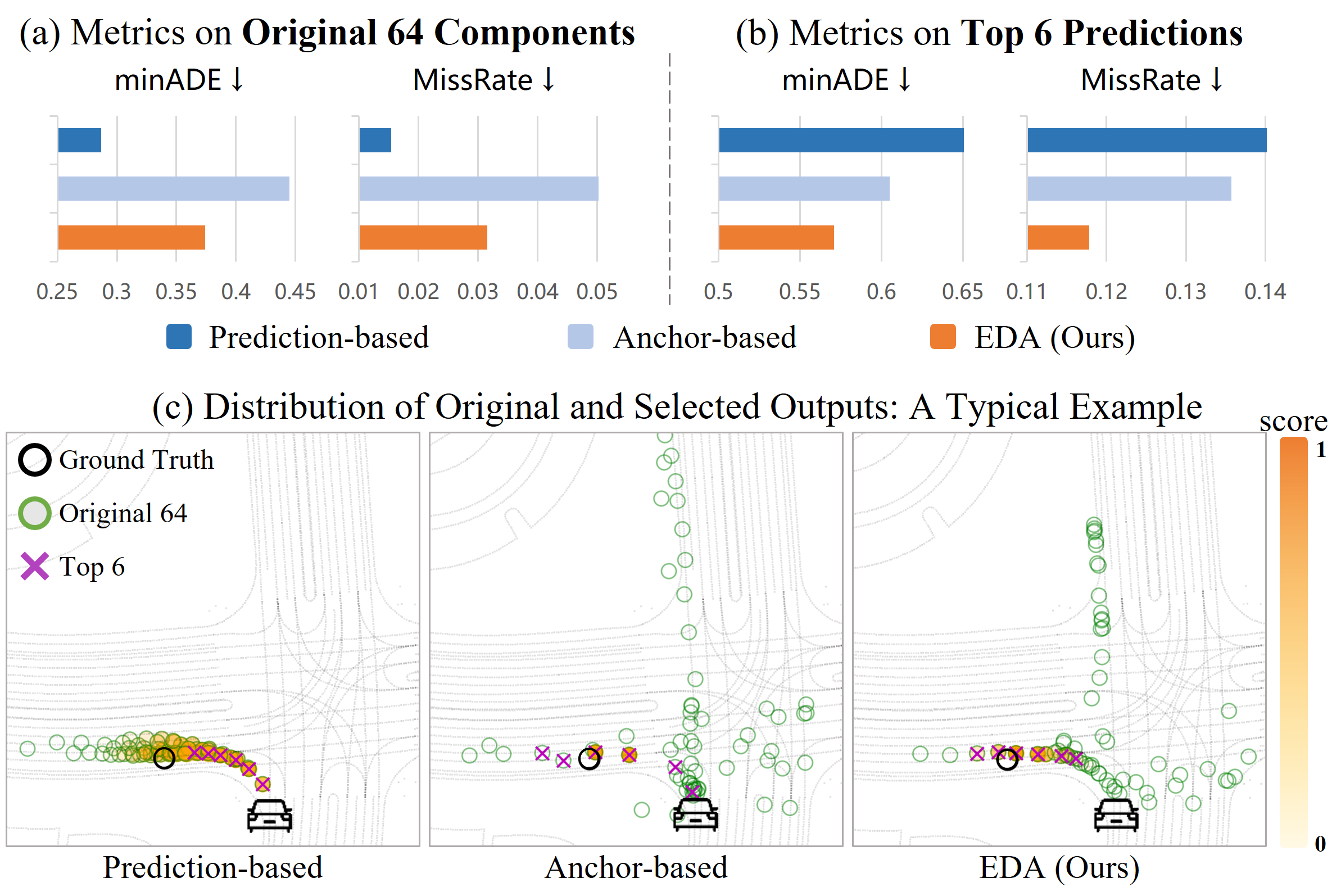} % Reduce the figure size so that it is slightly narrower than the column. Don't use precise values for figure width.This setup will avoid overfull boxes.
\caption{
The outcomes from different matching paradigms.
All of the strategies share the same network structure with 64 learnable queries.
The top 6 predictions are selected from the original ones by non-maximum suppression (NMS).
}
\label{fig:matching_results}
\end{figure}

\begin{figure*}[tb]
\centering
\includegraphics[width=0.95\textwidth]{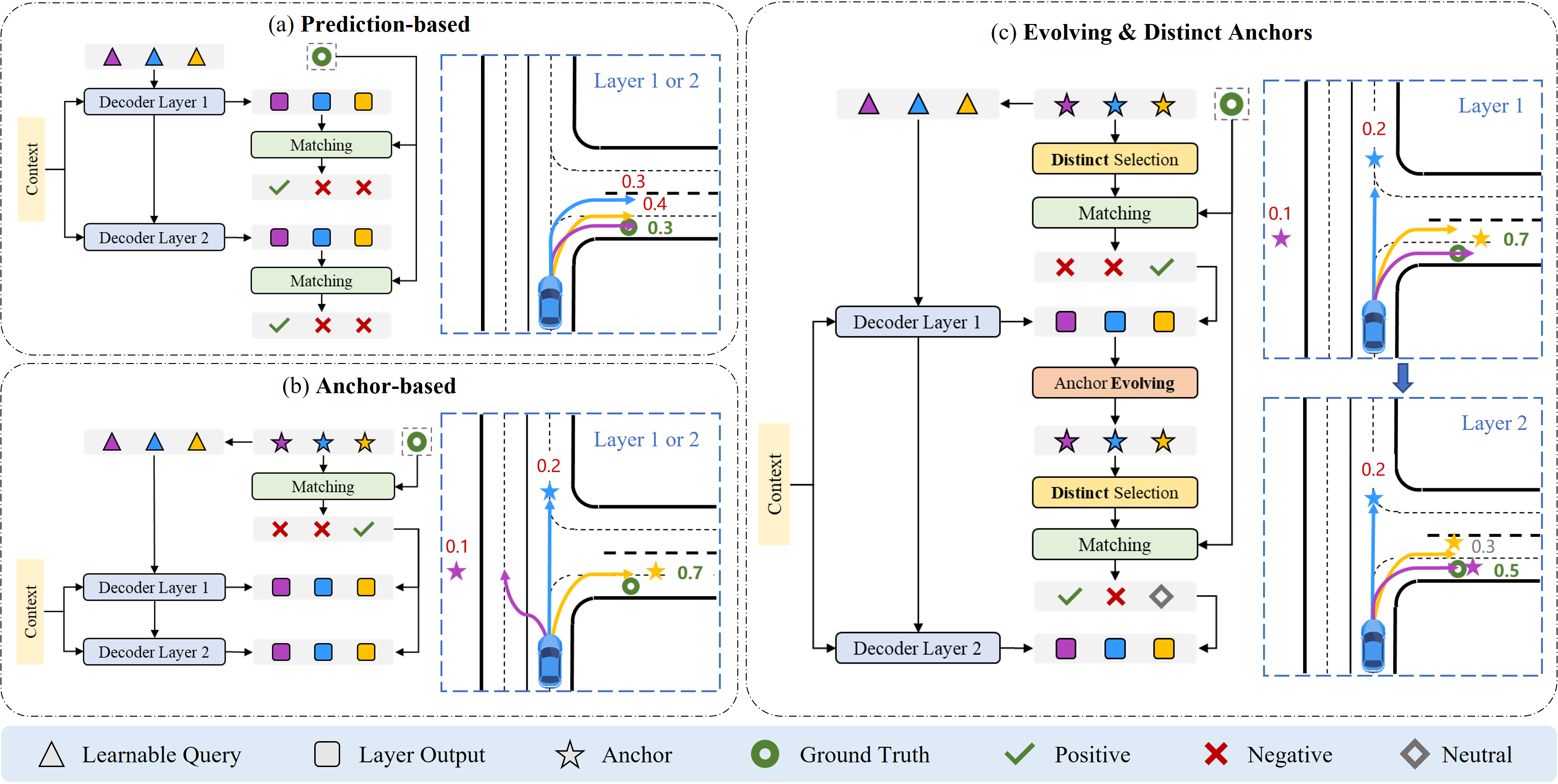} % Reduce the figure size so that it is slightly narrower than the column.
\caption{
The demonstration of different matching paradigms with a 2-layer decoder.
Each subfigure displays a workflow on the left and corresponding illustration on the right.
Objects with the same internal color belong to the same mixture component.
The numbers attached to each component represent the scores.
(a) and (b) respectively present the \textbf{prediction-based} and \textbf{anchor-based} matching.
(c) demonstrates the design of proposed \textbf{Evolving and Distinct Anchors (EDA)}, where the anchors for the \nth{2} layer are updated using the outputs from the \nth{1} layer.
Additionally, a selection of distinct anchors is applied before matching.
As a result, the yellow component in the \nth{2} layer is excluded since it is close to the purple one but has a lower score.
}
\label{fig:matching_paradigms}
\end{figure*}

In the field of autonomous driving, motion prediction is an important task which contributes to scene understanding and safe planning.
Motion prediction utilizes historical agent states and road maps to predict the future trajectories of traffic participants.
In recent years, an increasing amount of research works~\shortcite{shi2023mtr++, nayakanti2023wayformer, shi2022motion, zhou2022hivt, liu2021multimodal, ye2021tpcn, liang2020learning, chai2019multipath, gupta2018social, lee2017desire} have focused on motion prediction.
A major challenge of motion forecasting is the multimodality of future behaviors, which means an agent could carry out one of many underlying possibilities.

A bunch of works~\cite{ngiam2021scene, varadarajan2022multipath++, shi2022motion, chai2019multipath} have adopted mixture models, like Gaussian Mixture Model~(GMM), to represent multimodal future behaviors and have gained great success, where potential trajectories are modeled as scored components.
These approaches typically employ a winner-takes-all regression loss in conjunction with a classification term, which necessitates identifying the positive and negative mixture components.
For selecting positive components, there are two main categories of existing methods: \textbf{prediction-based} and \textbf{anchor-based} matching.

The prediction-based matching methods~\cite{ngiam2021scene, varadarajan2022multipath++} choose the predicted trajectory that is closest to the ground truth as the positive component, which is demonstrated in Fig.~\ref{fig:matching_paradigms}(a).
Predictions generated by these methods honestly reflect the high degree of uncertainty in future behaviors, which results in an originally lower minimum error and miss rate~(Fig.~\ref{fig:matching_results}(a)).
However, as illustrated in Fig.~\ref{fig:matching_results}(c), the output trajectories from prediction-based matching tend to cluster around the most probable regions and similar scores are made upon such predictions, making it difficult to pick representative trajectories for downstream tasks~(Fig.~\ref{fig:matching_results}(b)).

As demonstrated in Fig.~\ref{fig:matching_paradigms}(b), the anchor-based matching methods~\cite{shi2022motion, chai2019multipath} associate each component with an anchor endpoint or trajectory, and select the positive one corresponding to the closest predefined anchor to ground truth.
The introduction of spatial priors considerably alleviates the burden of optimization in classification, and the methods would prefer to generate trajectories around the predefined anchors.
Nevertheless, to reduce computational costs and prevent compromising the scoring performance~\cite{shi2022motion}, the anchors are usually distributed in a sparser manner compared to the outputs from prediction-based matching.
Hence the regression capability of model is limited, which is shown in Fig.~\ref{fig:matching_results}(a).

In this paper, we introduce a novel paradigm, named \textbf{Evolving and Distinct Anchors (EDA)}, to define the positive and negative components for multi-modal motion prediction based on mixture models.
As illustrated in Fig.~\ref{fig:matching_paradigms}(c), we first pre-define anchors and then update them by the intermediate outputs, hence the name \textbf{Evolving Anchors}.
On the one hand, we utilize spatial priors in the form of predefined anchors to alleviate the difficulties in trajectory scoring posed by prediction-based matching approaches.
On the other hand, we allow anchors to redistribute themselves based on predictions under specific scenes for a promoted regression capability compared to the vanilla anchor-based matching.
As the anchors evolve multiple times, we observe that the prediction clustering issue previously presented in prediction-based matching arises and becomes pronounced, which continues to bother the optimization in scoring trajectories.
In order to mitigate the ambiguity in classification caused by the gathering problem, inspired by Dense Distinct Query~\cite{zhang2023dense} for object detection, we select \textbf{Distinct Anchors} through non-maximum suppression (NMS) before matching them with the ground truth, as demonstrated in Fig.~\ref{fig:matching_paradigms}(c).
The adoption of distinct anchors also encourages the model to prioritize the most probable component among similar ones, facilitating the selection of representative predictions for downstream jobs.
It turns out that our method leverages the benefits of both anchor-based and prediction-based matching~(as shown in Fig.~\ref{fig:matching_results}), and achieves state-of-the-art performance on the Waymo Open Motion Dataset~\cite{ettinger2021large}.

Our contributions can be summarized as follows:
\begin{enumerate}
\item We propose the Evolving Anchors for multimodal motion prediction based on mixture models, where we pre-define spatial anchors and then update them by the intermediate outputs.
This novel strategy strikes a balance between the existing anchor-based and prediction-based matching approaches.
\item We adopt Distinct Anchors to address the ambiguity in classification induced by prediction clustering phenomena.
Employing NMS on anchors before matching them with the ground truth, we reduce the optimization difficulty in trajectory scoring and enhance the selection of representative predictions for subsequent tasks.
\item We have performed experiments on the Waymo Open Motion Dataset~\shortcite{ettinger2021large}.
With the assistance of Evolving and Distinct Anchors, our single model has surpassed the performance of previous ensemble-free approaches, 
% and is even superior to ensemble methods in some metrics
exhibiting improvements on all metrics compared to the baseline MTR~\cite{shi2022motion}, particularly with a significant relative reduction of 13.5\% in Miss Rate.
\end{enumerate}

\section{Related Work}
\subsection{Architectures for Motion Prediction}
In recent times, there has been a significant increase in the study of motion prediction owing to the rising interest in autonomous driving.
Motion prediction involves using the past agent states and road maps to forecast the future paths of traffic participants.
Early studies~\cite{chai2019multipath, casas2020spagnn, park2020diverse, gilles2021home, casas2021mp3} commonly rasterize the inputs into images and capture the contextual information through CNNs.
LaneGCN~\cite{liang2020learning} and LaneRCNN~\cite{zeng2021lanercnn} construct lane graphs to efficiently represent the topology of road maps.
Recent works~\cite{gu2021densetnt, varadarajan2022multipath++, shi2022motion} have widely adopted the VectorNet~\cite{gao2020vectornet} representation scheme, which regards the road maps as polylines.
As Transformers~\cite{vaswani2017attention} have gained popularity, an increasing number of studies~\cite{liu2021multimodal, ngiam2021scene, jia2023hdgt} have utilized the attention mechanism to encode scene context.
Encouraged by the successful application of DETR~\cite{carion2020end}, many Transformer-based models~\cite{girgis2021latent, varadarajan2022multipath++, nayakanti2023wayformer} have adopted learnable queries in decoder to generate multiple potential future trajectories.
In our study, we utilize the architecture presented in MTR~\cite{shi2022motion}, which is an advanced transformer framework incorporating a local attention based encoder and a decoder with intention queries.

\subsection{Modeling for Multimodal Future Motion}
Previous studies have investigated different approaches for modeling multimodal future behaviors.
Earlier generative models~\cite{lee2017desire, gupta2018social, rhinehart2018r2p2, rhinehart2019precog} generate a collection of samples to represent the distribution of future.
Many other works~\cite{chai2019multipath, mercat2020multi, ngiam2021scene} have utilized mixture models to parameterize multi-modal predictions, which mainly fall into two lines: prediction-based and anchor-based matching, as elaborated in introduction.
In prediction-based matching methods~\cite{ngiam2021scene, varadarajan2022multipath++, nayakanti2023wayformer}, the positive mixture component is chosen by directly comparing predicted trajectories to the ground truth.
Some models~\cite{tang2019multiple, girgis2021latent} using the loss based on EM algorithm can also be viewed as prediction-based matching when its KL term converges.
Due to the challenge of selecting representative future trajectories, these methods have opted to use well-designed aggregation techniques~\cite{varadarajan2022multipath++, nayakanti2023wayformer}, or to directly utilize an end-to-end version~\cite{ngiam2021scene, girgis2021latent}.
However, their scoring performance still lags behind that of anchor-based matching methods.
The anchor-based matching~\cite{chai2019multipath, zhao2021tnt} regards as positive the component matching the closest predefined anchor to ground truth.
The HOME series~\cite{gilles2021home, gilles2022gohome} and DenseTNT~\cite{gu2021densetnt} can be considered as variations of anchor-based matching, where the anchors are the grids in heatmaps or target candidates placed on roads, but they require an additional sampling process to obtain the final predictions.
The MTR~\cite{shi2022motion} achieves remarkable scoring performance using predefined anchors, while its end-to-end prediction-based matching version demonstrates significantly better performance in terms of minimum error and miss rate.
Motivated by the findings, we propose a novel matching paradigm to exploit the regression potential hidden by the state-of-the-art anchor-based matching strategy.

\subsection{Dense Distinct Query for Label Assignment}
According to \citeauthor{zhang2023dense}, considering one-to-one label assignment in object detection, sparse queries cannot ensure a high recall, while dense queries inevitably bring more similar queries and face optimization challenges in classification.
Therefore, they propose Dense Distinct Queries (DDQ), in which dense queries are first laid and then distinct queries are selected for one-to-one assignments.
Inspired by DDQ~\cite{zhang2023dense}, we adopt distinct anchors to mitigate the ambiguity in trajectory scoring induced by prediction clustering phenomena.

\section{Evolving and Distinct Anchors}

\begin{figure*}[tb]
\centering
\includegraphics[width=0.93\textwidth]{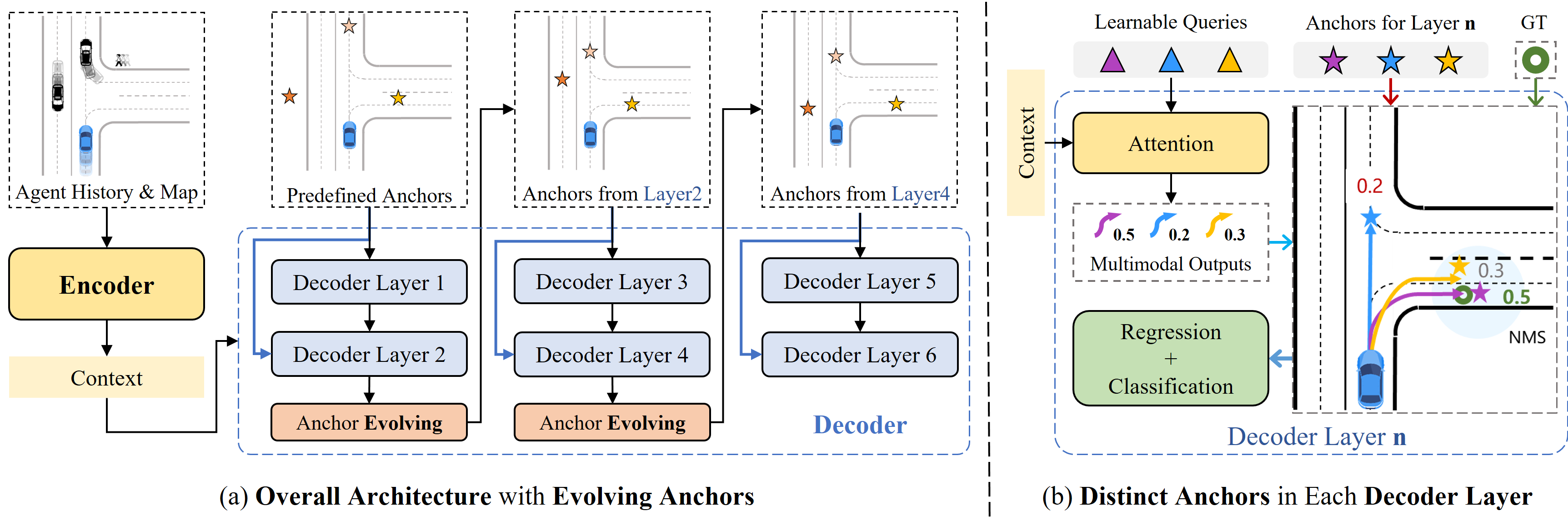} % Reduce the figure size so that it is slightly narrower than the column. Don't use precise values for figure width.This setup will avoid overfull boxes.
\caption{
The illustration of the EDA paradigm.
(a) shows an instance of the overall architecture with a 6-layer decoder and anchors evolving at the \nth{2}, \nth{4} layers.
(b) reveals the details in each decoder layer, where distinct anchors are selected before matching.
Components that correspond to the excluded anchors, such as the yellow one in picture, are considered neutral.
}
\label{fig:overall_architecture}
\end{figure*}

For identifying positive components, there are two primary strategies within the existing mixture-model based methods.
The \textit{prediction-based} matching directly compares the predicted trajectories $\{P_i\}_{i=1}^{N_C}$ with the ground truth $G$:
\begin{equation}
Distance(P_i, G), \ i = 1, \cdots, N_C,
\end{equation}
where $N_C$ denotes the number of components.
In \textit{anchor-based} matching, the spatial anchors $\{A_i\}_{i=1}^{N_C}$ are linked to each component and matched with the ground truth $G$:
\begin{equation}
Distance(A_i, G), \ i = 1, \cdots, N_C.
\end{equation}

In this study, we present \textit{Evolving and Distinct Anchors (EDA)}, a novel paradigm to define the positive and negative mixture components by:
\begin{equation}
Distance(A_{E_j}, G), \ j \in \mathcal{I}_D,
\end{equation}
where $A_E$ denotes the evolving anchors, and $\mathcal{I}_D$ is the index set of distinct anchors.
The main idea is illustrated in Fig.~\ref{fig:overall_architecture}.
In the following we first introduce the encoder-decoder structure upon which our method is built.
Subsequently, we provide detailed descriptions of the proposed \textit{Evolving Anchors} and \textit{Distinct Anchors} respectively.

\subsection{Network Architecture}
We have implemented our ideas on a cutting-edge encoder-decoder structure, as the one presented in MTR~\cite{shi2022motion}.
This transformer framework employs an encoder with local self-attention for scene context modeling, in addition to a multi-layer decoder that incorporates learnable intention queries to predict multimodal trajectories.

It is important to note that our approach presented in this paper is centered on the design of loss.
Consequently, the proposed \textit{Evolving and Distinct Anchors (EDA)} can be readily applied to any network structure that includes a multi-layer decoder.

\subsection{Evolving Anchors}
Although the spatial priors significantly alleviate the challenge in classification optimization, the vanilla anchor-based matching encounters a limitation in its regression capability, which will be demonstrated later.
Regarding the above issue and encouraged by the successful adoption of multi-layer decoders in motion prediction~\shortcite{girgis2021latent, shi2022motion}, we naturally consider enabling anchors to evolve through multiple decoder layers for an enlarged regression capacity.

Take a 6-layer decoder for instance, as illustrated in Fig.~\ref{fig:overall_architecture}(a), we can implement twice-evolving anchors by updating the anchors with outputs from the \nth{2} and \nth{4} layers, in which the evolving anchors for the $n$-th layer are:
\begin{equation}
A_E^{(n)} = \left\{
\begin{aligned}
    & A , & n=1,2 \\
    & P^{(2)} , & n=3,4 \\
    & P^{(4)} , & n=5,6 \\
\end{aligned}
\right.
,
\end{equation}
where we have omitted the index subscripts for simplicity.

In a word, the evolving anchors are \textit{initially predefined} and \textit{later adjusted} by the intermediate outputs from decoder layers, which means the anchors are allowed to redistribute themselves under specific scenes.

\begin{figure}[hbt]
\centering
\includegraphics[width=0.9\columnwidth]{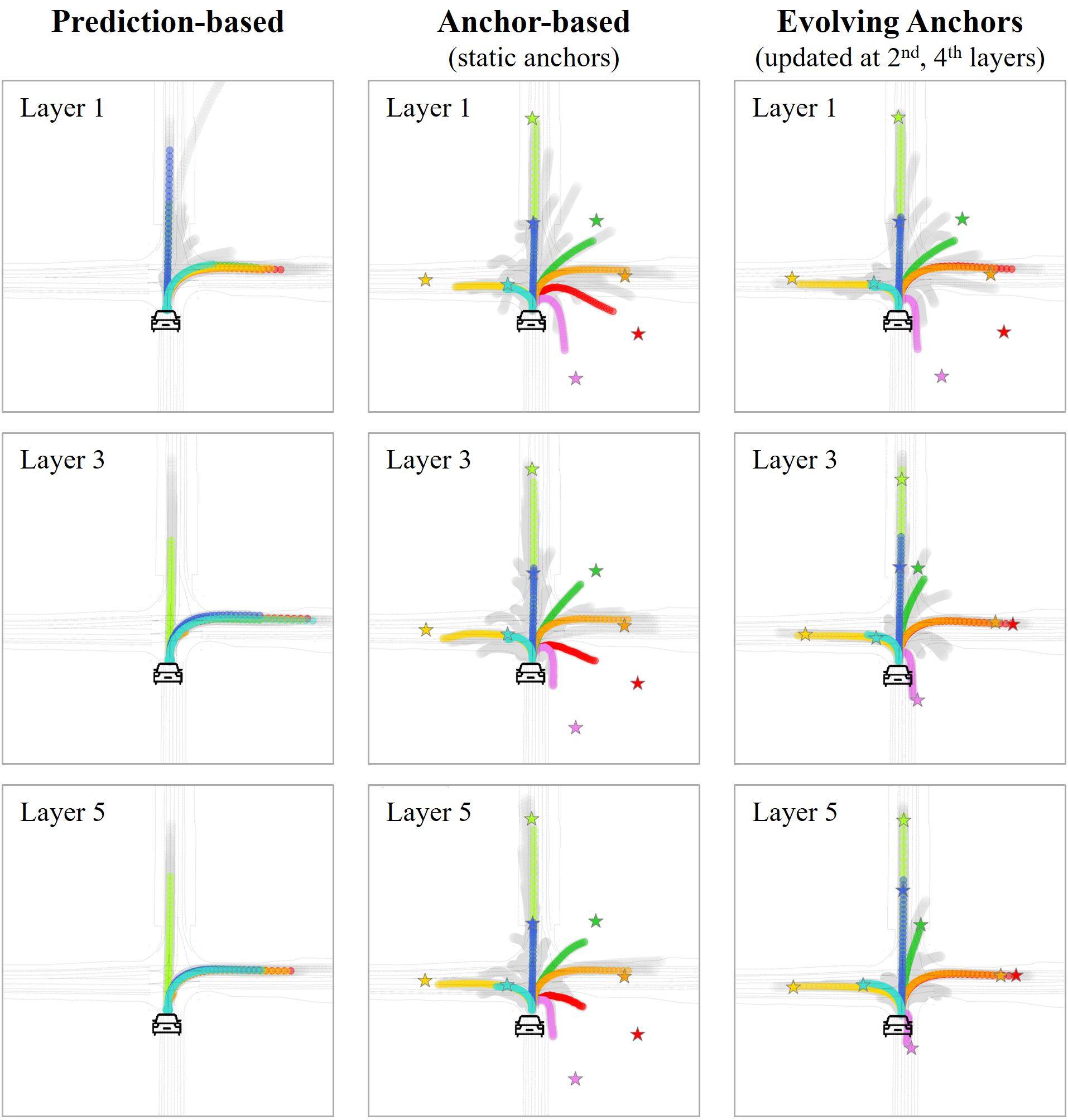} % Reduce the figure size so that it is slightly narrower than the column. Don't use precise values for figure width.This setup will avoid overfull boxes.
\caption{
Layer outputs from different matching paradigms under the same scene.
The $\star$ represents the anchor endpoint.
The typical trajectories are highlighted in bright colors, with each color indicating the same component across various methods, whereas the remaining ones are displayed in gray.
}
\label{fig:effects_of_anchors}
\end{figure}

\subsubsection{Effects of Evolving Anchors.}
The vanilla anchor-based matching, as presented in Fig.~\ref{fig:effects_of_anchors}, tends to make relative small adjustments to the predefined anchors in each layer.
This is because, making significant changes to the anchor that hits the ground truth would result in a considerable regression loss, while the refinements to unlikely ones are not encouraged.
Besides, the anchors are usually distributed in a sparser manner to reduce computational costs and avoid compromising the scoring performance~\cite{shi2022motion}.
Therefore, the regression capability of model is limited by the anchor-based matching with static anchors.

Correspondingly, making anchors adjustable motivates the model to modify unreasonable components in a larger degree, as illustrated in Fig.~\ref{fig:effects_of_anchors}.
Nevertheless, substantial refinements are made only when the potential benefits of achieving successful regression outweigh the expected cost of mistakenly making substantial adjustments.
Hence the modifications to anchors are restrained and progressive in evolving anchors.
In contrast, without the constraints from predefined anchors, the prediction-based matching would generate trajectories gathering around the most possible regions, even in the earlier layers, as shown in Fig.~\ref{fig:effects_of_anchors}.

Therefore, the proposed \textit{Evolving Anchors} achieves a balance between the anchor-based and prediction-based matching, where one can adjust the extent of modifications to predefined anchors through the frequency of anchor updates.

\subsection{Distinct Anchors}
Although predicting trajectories that cluster around the most probable regions contributes to better coverage of future behaviors with high uncertainty in prediction-based matching, this preference also introduces a serious issue of ambiguity in the scoring task.
With multiple gathering outcomes, it becomes difficult for the model to distinguish the actual one closest to the ground truth.
Hence the model tends to output similar scores for such predictions, making it hard to pick representative trajectories for downstream tasks.

\begin{figure}[thb]
\centering
\includegraphics[width=0.95\columnwidth]{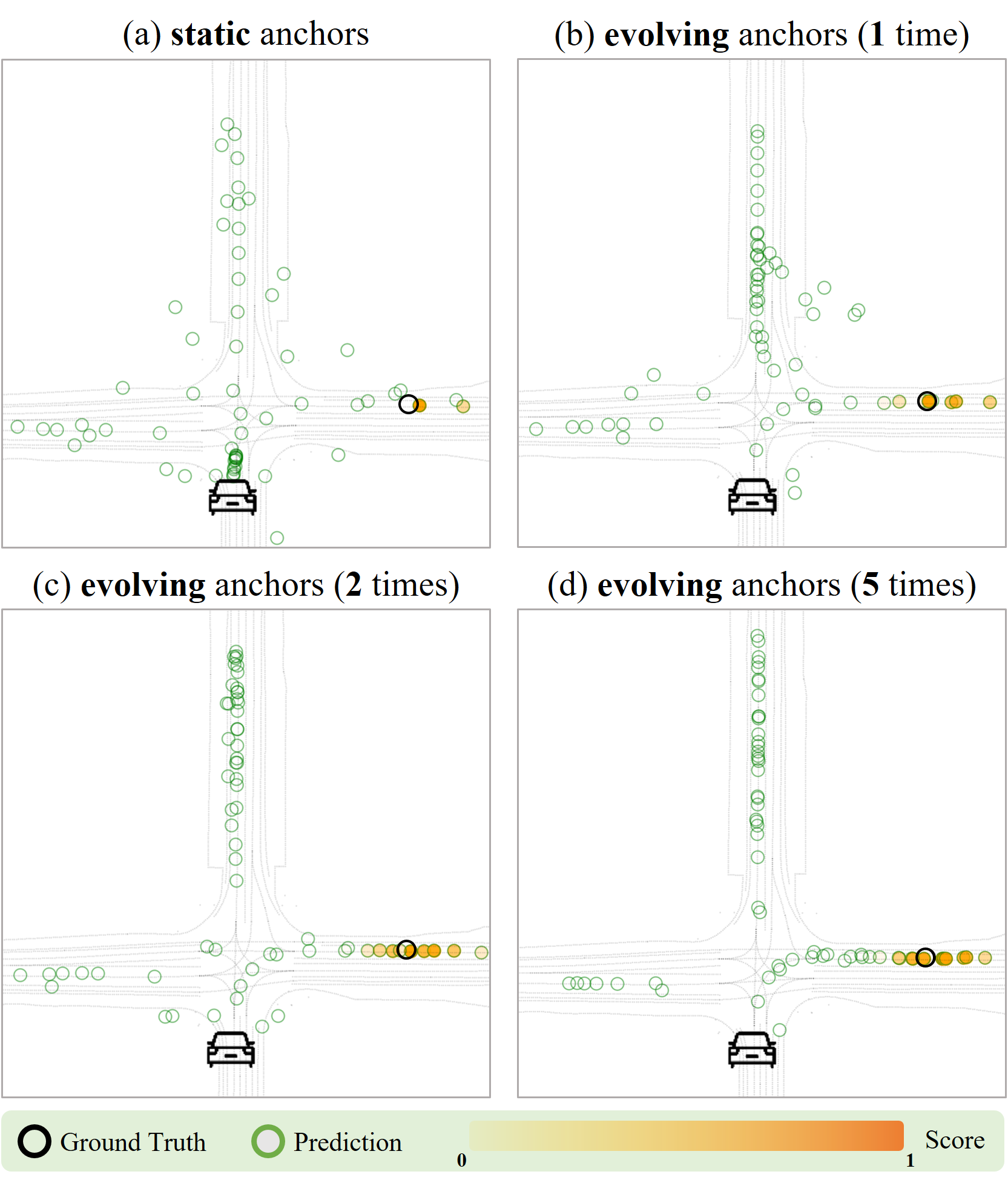} % Reduce the figure size so that it is slightly narrower than the column. Don't use precise values for figure width.This setup will avoid overfull boxes.
\caption{
A typical example illustrating the prediction clustering phenomenon in evolving anchors.
}
\label{fig:gathering_phenomena}
\end{figure}

In our proposed evolving anchors, as stated in the above analysis on \textit{effects of evolving anchors}, the more frequently we update anchors, the greater the opportunity for substantial adjustments to unreal components.
However, this also increases the potential for the phenomenon of prediction clustering.
Such patterns can be observed intuitively in Fig.~\ref{fig:gathering_phenomena}.
As a result, this issue continues to pose a challenge for optimization in classification, particularly when updating the anchors multiple times.

Taking inspiration from DDQ~\cite{zhang2023dense} in the object detection domain, we attempt to adopt distinct anchors to improve scoring performance.
Specifically, we apply non-maximum suppression (NMS) to the anchors for each decoder layer prior to matching them with the ground truth during training, as illustrated in Fig.~\ref{fig:overall_architecture}(b).
Mixture components that correspond to the excluded anchors will neither serve as positive nor negative samples.
Through the aforementioned operations:
\begin{itemize}
\item We prevent the labeling of similar anchors as opposite, which significantly reduces the optimization difficulty for the classification task.
\item Moreover, the model is encouraged to prioritize the most probable trajectory among the similar ones, making it easier to select the representative predictions using simple post-processing techniques such as NMS.
\end{itemize}

\subsection{Training Losses}
We train the model with a combination of winner-takes-all regression loss and classification term, which is commonly used in mixture-model based methods~\cite{chai2019multipath, nayakanti2023wayformer}.
Same as MTR~\cite{shi2022motion}, we employ a Gaussian regression loss.
Instead of Cross Entropy (CE) in MTR, we use Binary Cross Entropy (BCE) for classification loss, which is suitable for arbitrary numbers of mixture components filtered by distinct anchors.
Please refer to the Appendix for more implementation details.

\begin{table*}[htb]
\centering
\caption{
Top 6 metrics on the validation set of Waymo Open Motion Dataset~\cite{ettinger2021large}.
The terms ``original'', ``scaled'' and ``rank'' under the ``mAP'' heading respectively represent the results upon the original, scaled and ranking-oriented top 6 scores, as elaborated in implementation details.
}
\resizebox{0.85\textwidth}{!}{
\renewcommand{\arraystretch}{1.3}
\begin{tabular}{c|cc|cccccc}
\specialrule{2pt}{0pt}{1pt}
\multirow{2}{*}{\makecell{Anchor\\Evolving Times}} & \multirow{2}{*}{\makecell{Classification\\Loss}} & \multirow{2}{*}{\makecell{Distinct\\Anchors}} & \multicolumn{3}{c}{mAP $\uparrow$} & \multirow{2}{*}{minADE $\downarrow$} & \multirow{2}{*}{minFDE $\downarrow$} & \multirow{2}{*}{Miss Rate $\downarrow$} \\
\cline{4-6} &  &  & original & scaled & rank &  &  &  \\
\hline
\rowcolor{gray!20} 0 & CE &  & 0.4059 & 0.4167 & 0.4121 & 0.6012 & 1.2277 & 0.1348 \\
0 & BCE &  & 0.4053 & 0.4171 & 0.4126 & 0.6050 & 1.2376 & 0.1357 \\
\hline
\rowcolor{gray!20} 1 & CE &  & 0.4013 & 0.4211 & 0.4183 & 0.5867 & 1.2109 & 0.1240 \\
1 & BCE &  & 0.4060 & \textbf{0.4255} & 0.4228 & 0.5838 & 1.2012 & 0.1221 \\
1 & BCE & \checkmark & 0.4173 & 0.4221 & 0.4278 & 0.5776 & 1.1895 & 0.1203 \\
\hline
\rowcolor{gray!20} 2 & CE &  & 0.3868 & 0.4107 & 0.4101 & 0.5881 & 1.2145 & 0.1227 \\
2 & BCE &  & 0.3957 & 0.4236 & 0.4207 & 0.5888 & 1.2144 & 0.1229 \\
2 & BCE & \checkmark & \textbf{0.4235} & 0.4251 & \textbf{0.4353} & \textbf{0.5708} & \textbf{1.1730} & \textbf{0.1178} \\
\hline
\rowcolor{gray!20} 5 & CE &  & 0.3647 & 0.4051 & 0.4002 & 0.5996 & 1.2444 & 0.1264 \\
5 & BCE &  & 0.3675 & 0.4063 & 0.4037 & 0.5998 & 1.2412 & 0.1272 \\
5 & BCE & \checkmark & 0.4186 & 0.4185 & 0.4322 & 0.5817 & 1.2056 & 0.1245 \\
\specialrule{2pt}{0pt}{0pt}
\end{tabular}
}
\label{table:ablation}
\end{table*}

\begin{figure*}[htb]
\centering
\includegraphics[width=0.95\textwidth]{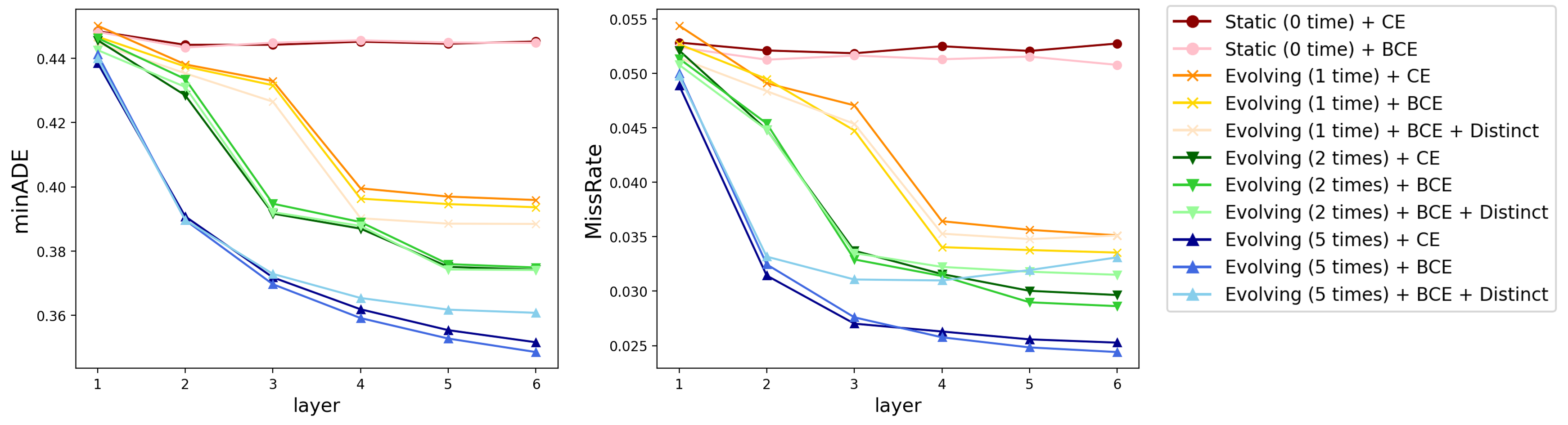} % Reduce the figure size so that it is slightly narrower than the column. Don't use precise values for figure width.This setup will avoid overfull boxes.
\caption{Minimum Error (left) and Miss Rate (right) on original 64 components for each decoder layer.}
\label{fig:original_results}
\end{figure*}

\section{Experiments}
\subsection{Experimental Setup}
\subsubsection{Dataset and metrics.}
We assess our method on the large-scale Waymo Open Motion Dataset (WOMD) proposed by \citeauthor{ettinger2021large}, which extracts interesting behaviors from actual traffic scenes.
The WOMD~\cite{ettinger2021large} includes 487k training scenes, 44k validation and 44k testing scenes, where each scene contains up to 8 target agents.
Each agent is comprised of 1 second of historical states and 8 seconds of future information.
The long time horizon challenges the model's capacity to capture a broad field of view and adapt to a vast output space for trajectories.

Due to the complexity of reasoning about numerous potential future behaviors, benchmark metrics limit the number of trajectories under consideration.
The official website offers an evaluation on submissions with up to 6 motion predictions for each target agent, returning metrics including minADE~(Minimum Average Displacement Error), minFDE~(Minimum Final Displacement Error), Miss Rate, Overlap Rate, mAP and Soft mAP.
Hence the top 6 metrics we provide are obtained from the official evaluation server, whereas we utilize a local evaluation tool based on the official API to compute metrics on a greater number of mixture components.

\subsubsection{Implementation details.}
Our design is built upon the state-of-the-art MTR framework \cite{shi2022motion}, where we adopt the default setting of the network structure and training configuration.
We train the model for 30 epochs on 16 GPUs (NVDIA RTX 3090) with the batch size of 80 scenes.
The predefined anchors we use are the 64 intention points generated by a k-means clustering algorithm on the training set, as used in MTR.
To achieve a more stable matching, except for predefined anchors we assign labels based on the full trajectories of intermediate outputs that act as evolving anchors.

For evaluation, we pick top 6 predictions by employing NMS on the endpoints of 64 predicted trajectories.
Following \citeauthor{shi2022mtra}, the distance threshold $\sigma$ is scaled proportionally to the length $L$ of trajectory with the highest confidence:
$\sigma = \min[3.5, \max[2.5, 2.5 + 1.5 \times (L-10)/(50-10)]]$.
The same NMS distance threshold is also applied to the selection of distinct anchors.
To improve the mAP metrics, the MTR~\shortcite{shi2022motion} scales the original top 6 scores for each sample through dividing them by their sum, making the scores comparable across different agents.
As far as we are concerned, it also makes sense to consider the rank of trajectories in a sample when comparing predictions across different agents.
Therefore, We add a rank-related integer to the original scores ranging between 0 and 1, to ensure that when computing the mAP metrics, the top-ranked trajectories of all samples will be sorted at the top, followed by the \nth{2}-ranked, \nth{3}-ranked, and so on.
For instance, we add 5 for the top-ranked trajectory, 4 for the \nth{2}-ranked, 3 for the \nth{3}-ranked, and so forth.
In order to align with previous works, we still present the mAP metrics upon the original and scaled scores in the following ablation study.

\begin{table*}[htb]
\centering
\caption{
Performance comparison on the validation and test sets of
Waymo Open Motion Dataset~\cite{ettinger2021large}.
% $\dag$: The results shown in \textit{italic} for reference is achieved with model ensemble techniques.
}
\resizebox{0.9\textwidth}{!}{
\renewcommand{\arraystretch}{1.3}
\begin{tabular}{c|c|cccccc}
\specialrule{2pt}{0pt}{1pt}
Set & Method & Soft mAP $\uparrow$ & mAP $\uparrow$ & minADE $\downarrow$ & minFDE $\downarrow$ & Miss Rate $\downarrow$ & Overlap Rate $\downarrow$ \\
\hline
\multirow{8}{*}{Test} & MotionCNN~\shortcite{konev2022motioncnn} & - & 0.2136 & 0.7400 & 1.4936 & 0.2091 & 0.1560\\
& ReCoAt~\shortcite{huang2022recoat} & - & 0.2711 & 0.7703 & 1.6668 & 0.2437 & 0.1642\\
& DenseTNT~\shortcite{gu2021densetnt} & - & 0.3281 & 1.0387 & 1.5514 & 0.1573 & 0.1779 \\
& SceneTransformer~\shortcite{ngiam2021scene} & - & 0.2788 & 0.6117 & 1.2116 & 0.1564 & 0.1473 \\
& HDGT~\shortcite{jia2023hdgt} & - & 0.2854 & 0.5933 & 1.2055 &  0.1511 & - \\
& MTR~\shortcite{shi2022motion} & 0.4216 & 0.4129 & 0.6050 & 1.2207 & 0.1351 & 0.1277 \\
& MTR++~\shortcite{shi2023mtr++} & 0.4414 & 0.4329 & 0.5906 & 1.1939 & 0.1298 & 0.1281 \\
& EDA (Ours) & \textbf{0.4510} & \textbf{0.4401} & \textbf{0.5718} & \textbf{1.1702} & \textbf{0.1169} & \textbf{0.1266} \\
% \cline{2-8}
% & $^\dag$Wayformer~\shortcite{nayakanti2023wayformer} & \textit{0.4335} & \textit{0.4190} & \textit{0.5454} & \textit{1.1280} & \textit{0.1228} & \textit{0.1270} \\
% & $^\dag$MTR++\_Ens~\shortcite{shi2023mtr++} & \textit{0.4738} & \textit{0.4634} & \textit{0.5581} & \textit{1.1166} & \textit{0.1122} & \textit{0.1276} \\
\hline
\multirow{3}{*}{Val} & MTR~\shortcite{shi2022motion} & - &  0.4164 & 0.6046 & 1.2251 & 0.1366 & - \\
& MTR++~\shortcite{shi2023mtr++} & - &  0.4351 & 0.5912 & 1.1986 & 0.1296 & - \\
& EDA (Ours) & \textbf{0.4462} & \textbf{0.4353} & \textbf{0.5708} & \textbf{1.1730} & \textbf{0.1178} & \textbf{0.1273} \\
\specialrule{2pt}{0pt}{0pt}
\end{tabular}
}
\label{table:benchmark}
\end{table*}

\subsection{Ablation Study}
We first investigate the impacts of \textit{Evolving Anchors}, and then assess the effectiveness of \textit{Distinct Anchors}.
All models are evaluated on the validation set of WOMD~\cite{ettinger2021large}.
In terms of mAP metrics, the results based on the original, scaled, and ranking-oriented top 6 scores are all presented, as referred in implementation details.

\subsubsection{Evolving Anchors.}
Starting from the baseline with 0 time of anchor updating, which is actually the MTR~\cite{shi2022motion} that uses the anchor-based matching with static anchors, we apply various \textit{anchor evolving times} to explore the effects of evolving anchors.
Upon the adopted 6-layer decoder, we update anchors at the \nth{3} layer for once-updating anchors, at the \nth{2} and \nth{4} layers for twice-evolving anchors, and at every but the final layer for 5 times of anchor evolving.
The corresponding top 6 metrics are displayed in the rows highlighted in gray of Table \ref{table:ablation}, while the results on original 64 components are included in Fig.~\ref{fig:original_results}.

Fig.~\ref{fig:original_results} shows that the regression capacity of model improves as the number of anchor updates increases, with a significant enhancement each time the anchors evolve.
This finding supports the idea that evolving anchors present opportunities to unlock the potential in regression hidden by the vanilla anchor-based matching.
And the more frequently we update the anchors, the greater the potential for adjustments to enhance the regression.

However, as illustrated in Fig.~\ref{fig:gathering_phenomena}, the phenomenon of prediction clustering also becomes severe when the anchors are updated more times, since the increased freedom in modifying the predefined anchors results in outputs more resembling those from the prediction-based matching.
This issue adversely affects the performance of trajectory scoring, leading to a decline in top 6 metrics when two or more anchor updates are employed, as presented in Table~\ref{table:ablation}.

\subsubsection{Distinct Anchors.}
We utilize the BCE loss to accommodate varying numbers of the mixture components selected for distinct anchors, which is different from the MTR~\cite{shi2022motion} using the CE loss.
Hence we begin by assessing the influence of various options for the classification loss.
From both Fig.~\ref{fig:original_results} and Table \ref{table:ablation}, it can be observed that, overall, the BCE loss leads to only marginal differences in the results, along with a slightly better mAP.
This suggests that the BCE loss can be considered a reasonable substitute for the CE classification loss.

After validating the impact of BCE loss, we now evaluate the efficacy of \textit{Distinct Anchors}.
As seen in Table \ref{table:ablation}, the use of distinct anchors brings a considerable enhancement in the top 6 metrics for models with evolving anchors.
What's more, the progress, particularly in mAP (\textit{e.g.}, +0.5\%, +1.46\%, +2.85\% for 1, 2, 5 anchor updates respectively upon ranking-oriented scores), becomes notable with a higher frequency of anchor evolving.
Nevertheless, the regression metrics on original 64 mixture components, as shown in Fig. \ref{fig:original_results}, do not exhibit a significant improvement.
Such evidences indicate that the adoption of distinct anchors does facilitate the selection for the representative behaviors as well as the scoring performance, which is hindered by the prediction clustering phenomenon.

But the benefits of distinct anchors are not limitless.
As depicted in Table~\ref{table:ablation}, both with the help of distinct anchors, the performance of 5 anchor updates cannot surpass that of twice-evolving anchors at all.
And the unusual deterioration in Miss Rate when using distinct anchors for 5 times of anchor evolving (Fig.~\ref{fig:original_results}) implies that the model may be still plagued by too many anchor updates.

\subsection{Benchmark Results}
We evaluate the model that performs the best in our ablation study, namely \textit{twice-evolving and distinct anchors} with ranking-oriented top 6 scores, on the test set of WOMD~\cite{ettinger2021large}.
We need to point out that the model for testing is trained solely on the WOMD training set without any ensemble techniques applied, consistent with our baseline MTR~\cite{shi2022motion}.

As shown in Table~\ref{table:benchmark}, our single model outperforms previous ensemble-free approaches on the WOMD.
The proposed EDA has demonstrated significant improvements in all performance metrics compared to the baseline MTR on both the validation and test sets.
Specifically, there is a relative improvement of 13.5\% on Miss Rate, 5.5\% on minADE, and 4.1\% on minFDE, as well as a +2.94\% absolute growth in SoftmAP on the test set.
Furthermroe, the performance of our EDA surpasses that of MTR++~\cite{shi2023mtr++}, the latest improved version of MTR, on both the validation and test sets of WOMD.
It is worth noting that MTR++ primarily enhances the network structure of MTR, while our approach centers on the design of loss, which means that combining the two complementary refinements has the potential to yield even more remarkable performance.
Please refer to the Appendix for more experimental results.

\section{Conclusions}
In this paper, we present Evolving and Distinct Anchors (EDA),  a novel paradigm to define the positive and negative components for multi-modal motion prediction based on mixture models.
We pre-define anchors and update them with intermediate outputs and pick distinct anchors before matching them with the ground truth.
Allowing the anchors to evolve and redistribute themselves under specific scenes promotes the regression capability of model.
The adoption of distinct anchors addresses the ambiguity in classification induced by the prediction clustering issue, and facilitates the selection of representative predictions for downstream tasks.
It turns out that our approach exhibits a significant improvement compared to the baseline MTR, achieving state-of-the-art performance on the Waymo Open Motion Dataset.

\newpage
\bibliography{aaai24}

% %File: appendix.tex

% title
\twocolumn[{
\begin{center}
{\fontsize{16pt}{32pt}\selectfont\textbf{Appendix}}
\end{center}
}]

\section{Implementation Details}
\subsection{Architecture details.}
\begin{figure*}[tb]
\centering
\includegraphics[width=0.8\textwidth]{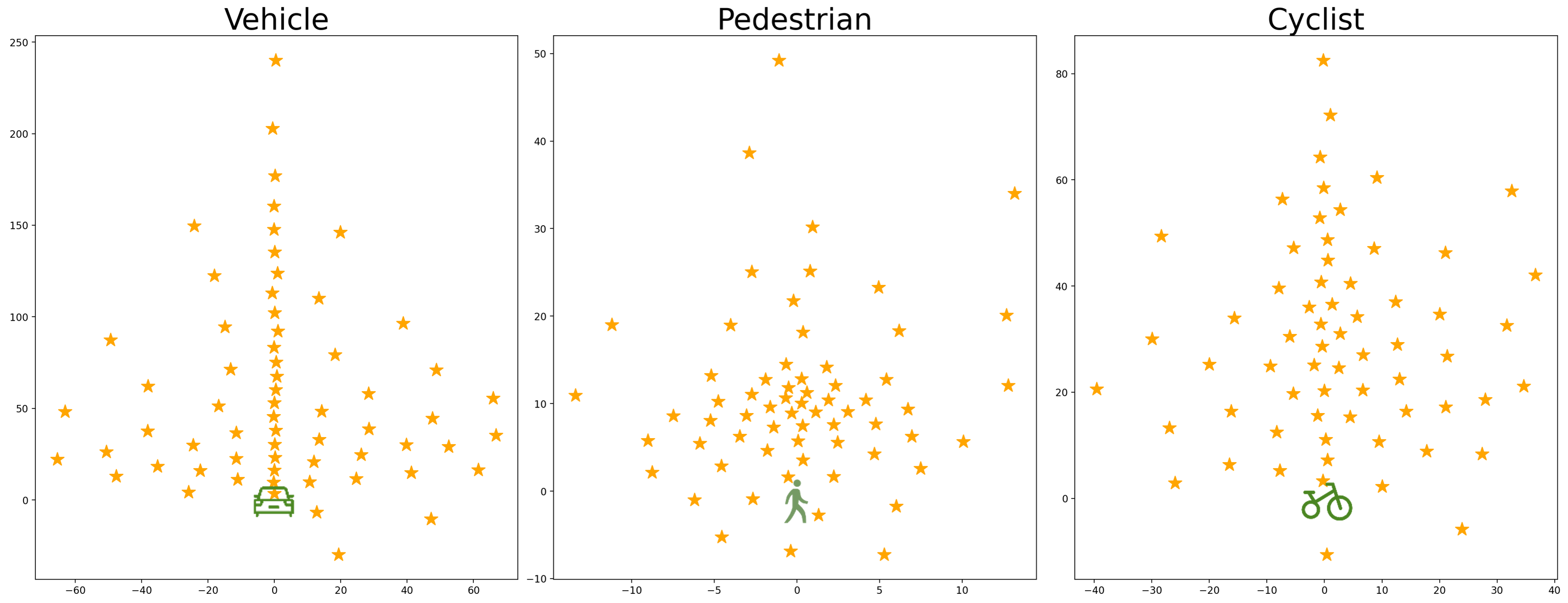} % Reduce the figure size so that it is slightly narrower than the column. Don't use precise values for figure width.This setup will avoid overfull boxes.
\captionof{figure}{
The distribution of predefined anchors shown as $\star$ for each type of traffic participants.
}
\label{fig:intention_points}
\end{figure*}

We have implemented our methods based on the Motion TRansformer~\cite{shi2022motion}.
To predict the motion of a target agent, we utilize the agent-centric strategy in which all inputs are normalized to the coordinate system centered around the agent in question.
The MTR adopts the vectorized representation~\cite{gao2020vectornet} to arrange the input agent states and road maps as polylines.
The encoder extracts the scene context information including agent and map features with local self-attention on the input polylines.
A dense future prediction generates the future states for all surrounding agents to boost their features.
Taking the context features as keys and values, a multi-layer decoder incorporates learnable intention queries to produce multimodal trajectories.
The learnable queries are initialized with intention endpoints generated by a k-means clustering algorithm on the training set.
The endpoints also serve as the predefined anchors for the vanilla anchor-based matching and our EDA, as shown in Fig.~\ref{fig:intention_points}.
In cross attention for each decoder layer, a dynamic map collection is applied to collect the closest map features to the latest predicted trajectories for querying, and the query position embedding is continuously updated using the layer outputs.
We adopt the default parameters of MTR in our experiments, which are shown in Table~\ref{table:architecture}.

\begin{table}[htb]
\centering
\caption{
The architecture parameters.
}
\renewcommand{\arraystretch}{1.2}
\begin{tabular}{c|lr}
\hline
\multirow{5}{*}{Enc.}
& number of layers & 6 \\
& number of map polylines & 768 \\
& number of points in each polyline & 20 \\
& number of attention neighbors & 16 \\
& hidden feature dimension & 256 \\
\hline
\multirow{3}{*}{Dec.}
& number of layers & 6 \\
& number of components & 64 \\
& number of nearest map polylines & 128 \\
\hline
\end{tabular}
\label{table:architecture}
\end{table}

\subsection{The mixture model loss.}
Given the context $X$ and the ground truth $Y$, the aim of optimization is to maximum the log-likelihood:
\begin{small}
\[
\begin{aligned}
&log \ p_{\theta}(Y|X)
= E_{Z\sim q(Z)}[log \frac{p_{\theta}(Y,Z|X)}{p_{\theta}(Z|Y,X)}] \\
= &E_{Z\sim q(Z)}[log \frac{p_{\theta}(Y,Z|X)}{q(Z)}] + E_{Z\sim q(Z)}[log \frac{q(Z)}{p_{\theta}(Z|Y,X)}] \\
= &E_{Z\sim q(Z)}[log \ p_{\theta}(Y,Z|X)] - H[q] + KL[q(Z)\|p_{\theta}(Z|Y,X)] \\
\ge &E_{Z\sim q(Z)}[log \ p_{\theta}(Y,Z|X)] - H[q] \\
= &E_{Z\sim q(Z)}[log \ p_{\theta}(Y|Z,X)] + E_{Z\sim q(Z)}[log \ p_{\theta}(Z|X)] - H[q] \\
= &E_{Z\sim q(Z)}[log \ p_{\theta}(Y|Z,X)] - KL[q(Z)\|p_{\theta}(Z|X)],
\end{aligned}
\]
\end{small}
where $\theta$ denotes the learnable weights of neural network, and $Z$ is the discrete random variable representing the probability of each component in \textit{mixture models}.
If we take:
\[
q(Z) = \mathds{1}(Z=z^*),
\]
where $z^*$ corresponds to the positive component whose anchor or prediction is closest to the ground truth, namely we utilize a winner-takes-all strategy.

Correspondingly, the term \begin{small}$E_{Z\sim q(Z)}[log \ p_{\theta}(Y|Z,X)]$\end{small} is the regression loss between the prediction of positive component $\hat{Y}^*$ and ground truth $Y$, where we model the probability using a Gaussian distribution:
\[
L_{reg} = log \ p_{\theta}(Y|Z=z^*,X) = log \ \mathcal{N}(Y|\hat{Y}^*).
\]
Since $H[q]$ is irrelevant to the learnable weights $\theta$, the other term \begin{small}$- KL[q(Z)\|p_{\theta}(Z|X)]$\end{small} is equivalent to the Cross Entropy loss on predicted scores $\hat{Z}$:
\[
\begin{aligned}
- KL[q(Z)\|p_{\theta}(Z|X)] &\Leftrightarrow  E_{Z\sim q(Z)}[log \ p_{\theta}(Z|X)] \\
&= -CE[\mathds{1}(Z=z^*), \hat{Z}].
\end{aligned}
\]
To accommodate varying numbers of the mixture components selected for distinct anchors, we employ Binary Cross Entropy for classification, which our experiments have shown to be a viable alternative to the CE loss:
\[
L_{cls} = -BCE[\mathds{1}(Z_d=z_d^*), \hat{Z}_d],
\]
where the subscript $d$ indicates the components corresponding to the distinct anchors, and the positive prediction $\hat{Y}^*$ in regression should also be replaced by $\hat{Y}_d^*$, which means that only the selected anchors have the chance to be positive.

In conclusion, the mixture model loss can be expressed as a combination of regression loss and a classification term:
\[
\mathcal{L}_\mathrm{mixture \ model} = \lambda_{reg} L_{reg} + \lambda_{cls} L_{cls}.
\]

\subsection{The details in EDA.}
The proposed \textit{Evolving and Distinct Anchors (EDA)} is centered on the design of loss, where the anchors for identification of positive components are updated by the layer outputs and selected to be distinct before matching. 
Therefore, the paradigm can be readily applied to any network structure that includes a multi-layer decoder.
When selecting distinct anchors through NMS, we utilize the scores from each layer to sort the corresponding anchors, since these scores actually represent the probability of the relevant anchors.
Below is the pseudo-code in PyTorch~\cite{paszke2019pytorch} demonstrating the EDA paradigm:
\vspace{0.5em}
\begin{lstlisting}
for layer_idx in range(num_decoder_layers):
    # predictions of the current layer
    pred_trajs, pred_scores = pred_list[layer_idx]
    # evolving anchors for the current layer
    anchor_trajs = evolving_anchors[layer_idx]
    # selection of distinct anchors
    distinct_mask = nms(anchor_trajs, pred_scores)
    # compare anchors with ground truth
    distance = compute_distance(anchor_trajs, gt)
    # compute loss
    loss = mixture_model_loss(
        pred_trajs, pred_scores, gt
        distinct_mask, distance
    )
\end{lstlisting}

\subsection{Training details.}
To align with the baseline, in addition to the mixture model loss, we also incorporate a loss for dense future prediction and an L1 loss on the agent velocity.
As shown in Table~\ref{table:training}, we adopt the same training configuration settings as in MTR~\cite{shi2022motion}, where we train a single model for all three categories without any data augmentation.

\begin{table}[htb]
\centering
\caption{
The training configuration.
}
\renewcommand{\arraystretch}{1.1}
\begin{tabular}{lr}
\hline
number of epochs & 30 \\
batch size & 80 \\
optimizer & AdamW \\
initial learning rate & 0.0001 \\
learning rate schedule & epoch 22,24,26,28 \\
learning rate decay factor & 0.5 \\
weight decay & 0.01 \\
attention dropout & 0.1 \\
regression weight & 1.0 \\
classification weight & 1.0 \\
dense regression weight & 1.0 \\
velocity regression weight & 0.5 \\
\hline
\end{tabular}
\label{table:training}
\vspace{-1em}
\end{table}

\section{Per-class Results of EDA}
We report the per-category performance of our EDA, which is the same model used in the benchmark results, on the Waymo Open Motion Dataset~\cite{ettinger2021large} for reference, as shown in Table~\ref{table:per_class}.

\begin{table}[htb]
\centering
\caption{
Per-class Performance on the validation and test sets of Waymo
Open Motion Dataset~\cite{ettinger2021large}.
% $\dag$: The results shown in \textit{italic} for reference is achieved with model ensemble techniques.
}
\resizebox{0.99\columnwidth}{!}{
\renewcommand{\arraystretch}{1.3}
\begin{tabular}{c|c|cccc}
\specialrule{2pt}{0pt}{1pt}
Set & Category & mAP $\uparrow$ & minADE $\downarrow$ & minFDE $\downarrow$ & Miss Rate $\downarrow$\\
\hline
\multirow{4}{*}{Test}
& Vehicle       &0.4807	&0.6808	&1.3921	&0.1164 \\
& Pedestrian    &0.4390	&0.3426	&0.7080	&0.0670 \\
& Cyclist       &0.4008	&0.6920	&1.4106	&0.1673 \\
& \textbf{Avg}  &0.4401	&0.5718	&1.1702	&0.1169 \\
\hline
\multirow{4}{*}{Val}
& Vehicle       &0.4810	&0.6820	&1.3990	&0.1176 \\
& Pedestrian    &0.4279	&0.3431	&0.7124	&0.0658 \\
& Cyclist       &0.3970	&0.6873	&1.4077	&0.1700 \\
& \textbf{Avg}  &0.4353	&0.5708	&1.1730	&0.1178 \\
\specialrule{2pt}{0pt}{0pt}
\end{tabular}
}
\label{table:per_class}
\end{table}

\section{Inference Consumption}
As a paradigm to define positive components for mixture-model-based methods, EDA theoretically should \textit{not add any burden to the model during inference}.
To empirically validate the viewpoint, we test our baseline MTR and the EDA with varying evolving frequencies on average consumption for single-scene inference on Waymo dataset on a 3090 GPU.
As shown in Table~\ref{table:consumption}, all models exhibit identical inference latency, GPU memory usage and computational costs.

\begin{table}[htb]
\centering
\caption{The inference consumption of EDA.}
\resizebox{0.9\columnwidth}{!}{
\renewcommand{\arraystretch}{1.2}
\begin{tabular}{c|cccc}
\specialrule{2pt}{0pt}{1pt}
Method & Evolving Times & Latency & Memory Cost & MACs \\
\hline
MTR
& - &62ms &1.83G &30.2G \\
\hline
\multirow{3}{*}{EDA}
& 1 &62ms &1.83G &30.2G \\
& 2 &62ms &1.83G &30.2G \\
& 5 &62ms &1.83G &30.2G \\
\specialrule{2pt}{0pt}{0pt}
\end{tabular}
}
\label{table:consumption}
\end{table}

\section{Combined Effects of EDA}
Our outstanding performance benefits from the combined effect of evolving anchors \textit{and} distinct anchors.
The former enhances the regression capacity but also leads to prediction clustering issues, which are then addressed by the latter.
To further confirm the above conclusions, we conduct an experiment on the effect of solely using distinct anchors.
As shown in Table~\ref{table:distinct_anchors}, merely distinct anchors bring little to no improvement, since the predefined anchors are already uniformly distributed.

\begin{table}[htb]
\centering
\caption{The performance of only applying distinct anchors.}
\resizebox{0.9\columnwidth}{!}{
\renewcommand{\arraystretch}{1.1}
\begin{tabular}{c|cccc}
\specialrule{2pt}{0pt}{1pt}
Distinct Anchors & mAP $\uparrow$ & minADE $\downarrow$ & minFDE $\downarrow$ & MR $\downarrow$ \\
\hline
 &0.4171 &0.6050 &1.2376 &0.1357 \\
\checkmark &0.4208 &0.6037 &1.2342 &0.1359 \\
\specialrule{2pt}{0pt}{0pt}
\end{tabular}
}
\label{table:distinct_anchors}
\end{table}

\section{Ablation Study on Predefined Anchors}
Our ablation experiments on different numbers of anchors further demonstrate the \textit{generalizability} of our method.
As shown in Table~\ref{table:ablation_anchor}, the EDA presents excellent performance improvement across different sets of predefined anchors.
We also observe that when there are fewer anchors, more times of evolving would be helpful to enhance the regression capability, which is consistent with our analysis of \textit{effects of evolving anchors}.

\begin{table}[htb]
\centering
\caption{The additional ablation on anchors.}
\resizebox{0.99\columnwidth}{!}{
\renewcommand{\arraystretch}{1.2}
\begin{tabular}{c|cc|cccc}
\specialrule{2pt}{0pt}{1pt}
\multirow{2}{*}{\makecell{Number of\\Anchors}} & 
\multirow{2}{*}{\makecell{Evolving\\Times}} & 
\multirow{2}{*}{\makecell{Distinct\\Anchors}} & 
\multirow{2}{*}{\makecell{mAP $\uparrow$}} & 
\multirow{2}{*}{\makecell{minADE $\downarrow$}} & 
\multirow{2}{*}{\makecell{minFDE $\downarrow$}} & 
\multirow{2}{*}{\makecell{MR $\downarrow$}} \\
& & & & & & \\
\hline
\multirow{4}{*}{16}
& 0 &            &0.4009 &0.6553 &1.4137 &0.1769 \\
& 1 & \checkmark &0.4138 &0.6038 &1.2750 &0.1483 \\
& 2 & \checkmark &\textbf{0.4207} &0.5886 &1.2299 &0.1378 \\
& 5 & \checkmark &0.4201 &\textbf{0.5833} &\textbf{1.2041} &\textbf{0.1365} \\
\hline
\multirow{3}{*}{100}
& 0 &            &0.4158 &0.6023 &1.2263 &0.1333 \\
& 1 & \checkmark &\textbf{0.4302} &\textbf{0.5789} &\textbf{1.1917} &\textbf{0.1216} \\
& 2 & \checkmark &0.4231 &0.5838 &1.2083 &0.1236 \\
\specialrule{2pt}{0pt}{0pt}
\end{tabular}
}
\label{table:ablation_anchor}
\end{table}

\section{Additional Results on Argoverse 2}
We conduct experiments on another large-scale dataset Argoverse 2~\cite{Argoverse2}, with the SOTA method QCNet~\cite{Zhou_2023_CVPR} as the baseline. As shown in Table~\ref{table:qcnet_av2}, our method achieves \textit{consistent improvements}.

Decoder of the original QCNet consists of two stages: prediction-based proposal generation and anchor-based refinement.
To better validate the improvements brought by our proposed matching paradigm, we first construct a pure anchor-based variant of QCNet with minimal modifications, and then apply EDA to this modified version.
To align with the consumption (comparable MACs in Table~\ref{table:qcnet_av2}), we use only 16 predefined anchor trajectories and employ NMS to select the top 6 predictions for evaluation.
For evaluation, apart from the metrics commonly used in Argoverse 2, we further calculate the mAP with a combined endpoint threshold of 2, 4, and 6m.

Table~\ref{table:qcnet_av2} presents the performance of pure anchor-based variant of QCNet~(\textbf{Anchor}), ours~(\textbf{Anchor + EDA}), and original QCNet~(\textbf{Original}) as reference.
Due to its limited refinement capacity and the sparsity of predefined anchors, the pure anchor-based variant of QCNet struggles to generate accurate predictions.
In contrast,  EDA significantly improves the prediction quality and achieves the \textit{best mAP} among all models.
This indicates that our method not only enhances the regression capability but also exhibits remarkable scoring performance.

\begin{table}[htb]
\centering
\caption{Top 6 metrics on the validation set of Argoverse 2.}
\resizebox{0.99\columnwidth}{!}{
\renewcommand{\arraystretch}{1.2}
\begin{tabular}{l|cccc|c}
\specialrule{2pt}{0pt}{1pt}
\textbf{QCNet} & mAP $\uparrow$ & b-minFDE $\downarrow$ & minADE $\downarrow$ & MR $\downarrow$ & MACs \\
\hline
Original &0.50 &1.86 &0.72 &0.15 &30G \\
\hline
Anchor &0.48 &3.04 &1.20 &0.51 &31G \\
Anchor + EDA & \textbf{0.58} &2.10 &0.81 &0.23 &31G \\
\specialrule{2pt}{0pt}{0pt}
\end{tabular}
}
\label{table:qcnet_av2}
\end{table}

\section{Qualitative Results}
\begin{figure*}[hb]
\centering
\includegraphics[width=0.99\textwidth]{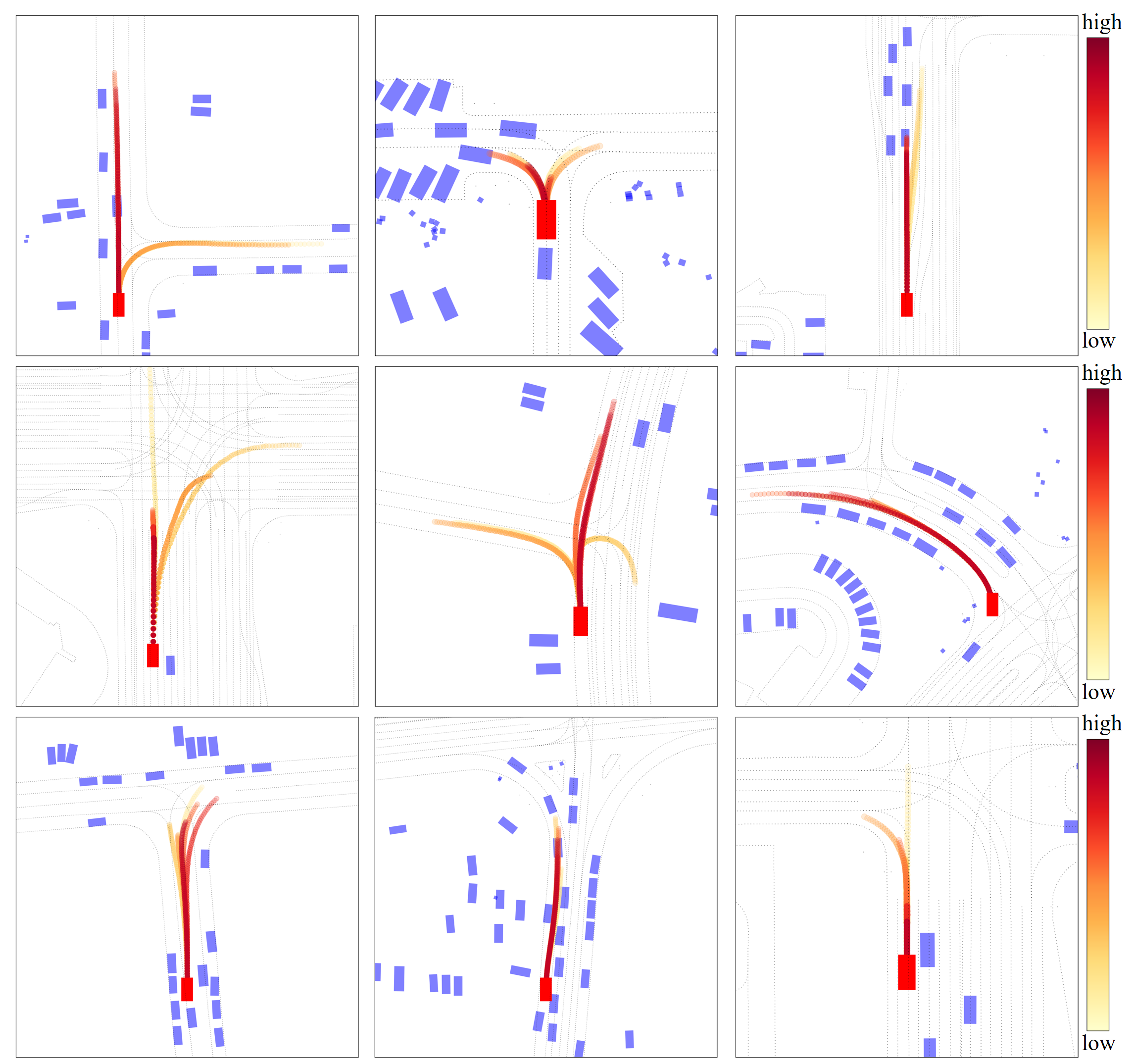} % Reduce the figure size so that it is slightly narrower than the column. Don't use precise values for figure width.This setup will avoid overfull boxes.
\caption{
Qualitative results of our EDA on the Waymo Open Motion Dataset~\cite{ettinger2021large}.
The top 6 predictions are displayed, with the darker colors indicating the higher scores.
The rectangular shapes indicate the current positions of the agents, where the red one is the target agent, and the blue ones represent the surrounding agents.
}
\label{fig:qualitative_results}
\end{figure*}

We provide some qualitative results of our EDA on the Waymo Open Motion Dataset~\cite{ettinger2021large} in Fig.~\ref{fig:qualitative_results}.

\end{document}